\documentclass{llncs}

\usepackage{graphicx, subfigure}
\usepackage{amssymb}
\usepackage{amsmath}
\usepackage{tabulary}
\usepackage{cite}
\usepackage{booktabs}
\begin{document}

\title{Enhanced Characterness for Text Detection in the Wild}
\titlerunning{Hamiltonian Mechanics}  
%
\author{Aarushi Agrawal\inst{2}, Prerana Mukherjee\inst{1}, Siddharth Srivastava\inst{1},   
        and Brejesh Lall\inst{1}}
\authorrunning{Aarushi Agrawal et al.} 
%
%
\institute{Department
of Electrical Engineering, Indian Institute of Technology, Hauz
Khas, Delhi, India \\
\email{\{eez138300, eez127506, brejesh\}@ee.iitd.ac.in},\\ 
\and
Dept of Electrical Engineering, Indian Institute of Technology  Kharagpur\\
\email{aarushiagrawal1995@gmail.com}
}

\maketitle              

\begin{abstract}
Text spotting is an interesting research problem as text may appear at any random place and may occur in various forms. Moreover, ability to detect text opens the horizons for improving many advanced computer vision problems. In this paper, we propose a novel language agnostic text detection method utilizing edge enhanced Maximally Stable Extremal Regions in natural scenes by defining strong characterness measures. We show that a simple combination of characterness cues help in rejecting the non text regions. These regions are further fine-tuned for rejecting the non-textual neighbor regions. Comprehensive evaluation of the proposed scheme shows that it provides comparative to better generalization performance to the traditional methods for this task.
\keywords{Text detection, HOG, enhanced MSER, stroke width}
\end{abstract}
\section{Introduction}
Text co-occurring in images and videos serve as a warehouse for valuable information for image description, thus assists in providing suitable annotations. Typical practical applications involve extracting street names and numbers, textual indications such as ``\textit{diversion ahead}'' etc. from road signs in natural scenes. Such information can be further stored in geo-tagged databases \cite{minetto2014snoopertext}. Autonomous vehicles are also heavily dependent on efficiency and accuracy of such methods to effectively follow traffic rules. Another area where text detection is applied is indexing and tagging images/videos where text in images helps in better understanding of the content \cite{ye2015text}. Performing the above tasks is trivial for humans but segregating it against a challenging background still remains as a complicated task for machines. Traditional methods for text detection employ the use of blob detection schemes like Maximally Stable Extremal Regions (MSERs) \cite{chen2011robust, neumann2012real}, edge based analysis, Stroke Width Transform (SWT) \cite{huang2013text, yi2012localizing}, strokelets \cite{yao2014strokelets} and features like Histogram of Oriented Gradients (HOG) \cite{minetto2014snoopertext, hanif2009text}, Gabor based features \cite{yi2012localizing}, text covariance descriptors \cite{huang2013text, sivic2003video} and shape descriptors (e.g. Fourier descriptors \cite{de2014automated, fabrizio2013text}, Zernike moments \cite{kan2002invariant}). The reason behind great popularity of using MSERs and SWT is their $O(n)$ time complexity for performing efficient segmentation which helps in detecting the text regions. MSERs are very effective in detecting the text components but it are extremely sensitive to noise. So, most of the techniques concentrate on pruning the non-text regions using some heuristics or geometric properties. Despite the advent of deep learning based techniques \cite{jaderberg2016reading, he2016text} which have resulted in tremendous progress in machine driven text detection, the traditional methods still hold relevance primarily owing to their simplicity and comparable generalization capability to different languages. 

Authors in \cite{li2014characterness} utilize text specific saliency detection measure termed as \textit{characterness}. The authors demonstrate that due to presence of contrasting objects, saliency alone cannot be an effective indicator of textual region. They overcome this limitation by introducing saliency cues which accentuate the boundary information in addition to saliency \cite{mukherjee2015saliency}. Deriving motivation from this work, we propose a simple combination of various characterness cues 
for generating candidate bounding boxes for text regions. We use these characterness cues (HOG, stroke width variance, pyramid histogram of oriented gradients (PHOG)) to refine the blobs generated by edge enhanced MSERs (eMSERs) \cite{li2014characterness} for generating text candidates. This is followed by rejection of non-text regions by incorporating difference of entropy as a discriminating factor. The last step is the refinement step, where we combine the smaller blobs into one single text region by concatenating blobs with similar stroke width variance and characterness cue distribution. 
As per the above discussion, the key contributions of the paper are listed below:
\begin{enumerate}
\item We develop a language agnostic text identification framework using text candidates obtained from edge based MSERs and  combination of various characterness cues. This is followed by a entropy assisted non-text region rejection strategy. Finally, the blobs are refined by combining regions with similar stroke width variance and distribution of characterness cues in respective regions. 
\item We provide comprehensive evaluation on popular text datases against recent text detection techniques and show that the proposed technique provides equivalent or better results.
\end{enumerate}

Organization of the paper is as follows: The proposed methodology is discussed in Section \ref{sec:Methodology}. The experimental analysis and the results is detailed in Section \ref{sec:results}. Finally the paper is concluded in Section \ref{sec:conclusion}.

\section{Proposed Methodology}\label{sec:Methodology}
\begin{figure}[hbtp]
\centering
{
\includegraphics[scale=0.45]{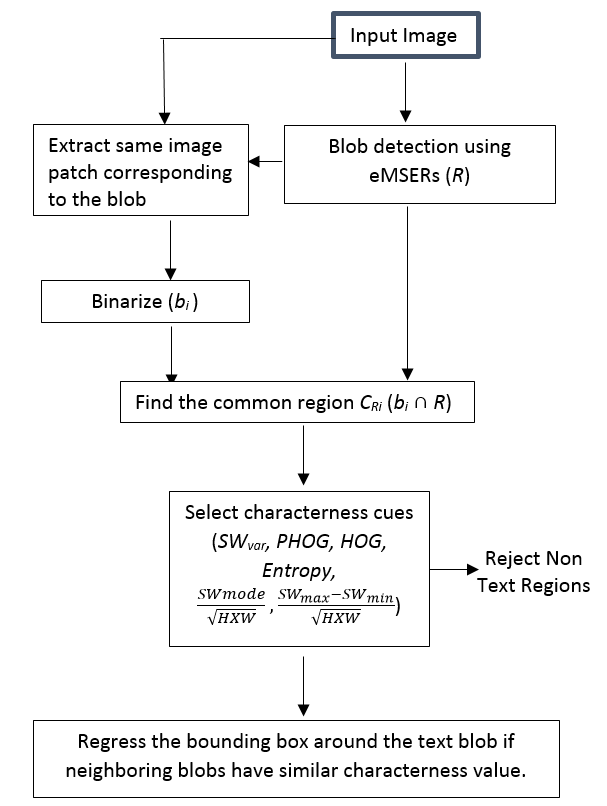}
\caption{Workflow of the proposed methodology}
\label{fig:block-diagram}
}
\end{figure}

The workflow of the proposed method is shown in Fig. \ref{fig:block-diagram}. In the following subsections, we describe in detail the components of the proposed method.

\subsection{Text candidate generation using eMSERs}
\begin{figure*}[ht]
\centering
\fbox
{
\includegraphics[scale=0.38]{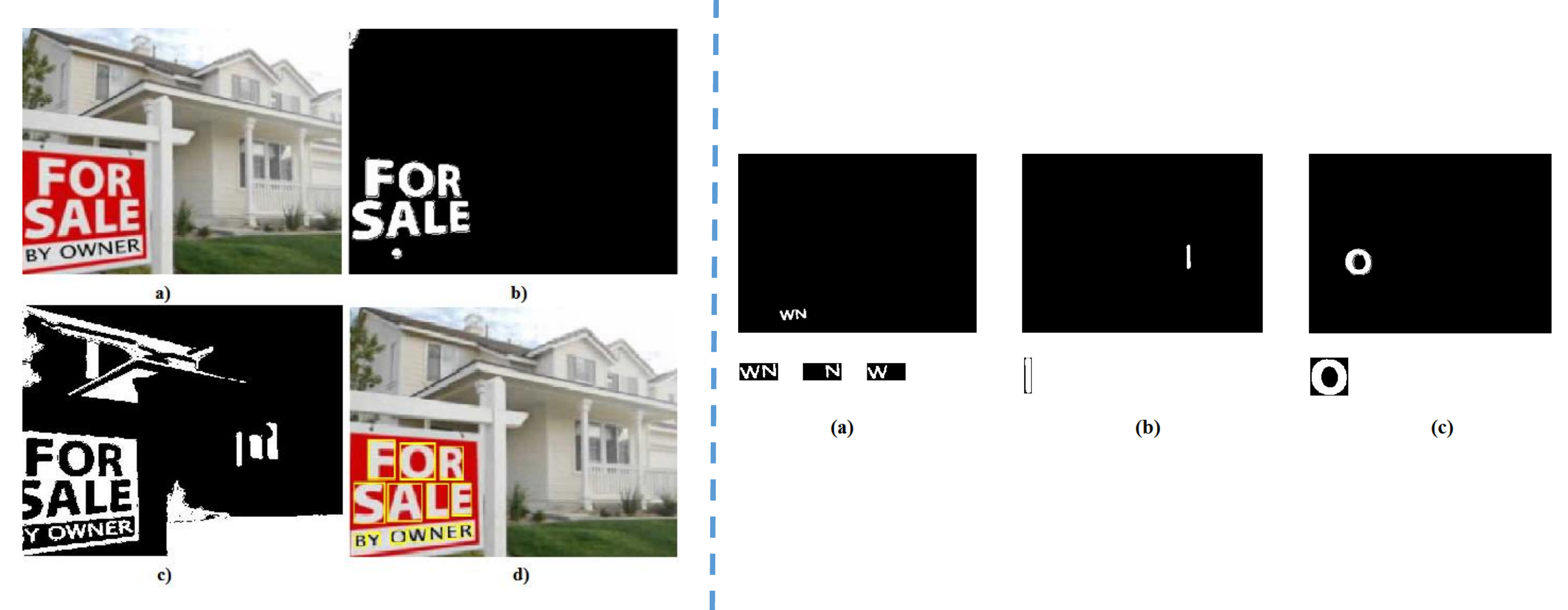}}
\caption{Left Column: (a) Original image (b) Bright regions (c) Dark regions (d) After processing on these regions final set of blobs detected by eMSERs; Right Column: Top Row: (a)-(c) eMSER region; Bottom Row: (a) Binarized region obtained from original image (b) Binarized region neglected due to size constraints (c) binarized image- refined object (alphabet) obtained with less disturbance which gives us better results}
\label{fig:emserdetect}

\end{figure*}

We begin by generating initial set of text candidates using edge enhanced Maximally Stable Extremal Regions (eMSERs) approach \cite{li2014characterness}. MSER is a method for blob detection which extracts the covariant regions in the image. It is based on aggregation of regions 
which have similar intensity values at various thresholds which makes it a suitable candidate for detecting regions with text in images. It efficiently detects the characters in case of distinctive boundaries but fails in the presence of blur. In order to handle this, eMSERS are computed over the gradient amplitude based image. It divides the image into two sets of regions: dark and bright; dark regions are those which have lower intensity than their surroundings and vice-versa. Initially non text regions are rejected based on geometric properties like aspect ratio, number of pixels and skeleton length followed by connected component analysis for combining the text regions. Fig. \ref{fig:emserdetect} shows instances of bright and dark regions formed during text candidate generation using eMSERs. As can be observed, in the bright regions the color of the text is lighter as compared to dark background (red) while in the dark regions the dark text was highlighted against the light colored background.

\subsection{Elimination of non-text regions}
The regions are further refined based on the property that text usually appears on a surrounding having a distinctive intensity. Utilizing this property we refine textual regions while reject non-textual regions. To achieve this, we find corresponding image patches for the blobs identified by eMSERs. As the image patches contain spurious data along with the information in the form of text, we perform binarization over these image patches using Otsu's threshold \cite{1975threshold} for that region and obtain a common region, $C_{Ri}$ between the binarized image patch $b_{i}$ and the blob obtained by eMSER $R$ (where $b_{i} \cap R$ $> 90\%$) for image $i$. A blob is rejected, if it is not contained in the binarized image patch. Fig. \ref{fig:emserdetect} shows some examples of this rejection strategy. We then define various characterness cues \cite{li2014characterness} for common regions $C_{Ri}$. Apart from stroke width and HOG used in \cite{li2014characterness}, we check the values of  pyramid histogram of oriented gradients (PHOG) features and entropy for the blobs. During the experiments we found that PHOG is a good measure of similarity over HOG. In case of alphabets i.e textual regions, we observe that the HOG and PHOG values for $C_{Ri}$ are very less. We now briefly explain these cues,
\begin{enumerate}
\item Stroke width variance: A stroke is effectively a continuous band of same width in an image. Stroke width transform (SWT) \cite{epshtein2010detecting} is defined as a local operator which gives the most likely stroke for every pixel in the image. In SWT, all the pixels are initialized with infinity as their stroke width. A Canny based edgemap is then calculated followed by calculation of gradient direction for all the edge pixels. If the gradient direction ($g_p$) of an edge pixel $p$ is opposite to the gradient direction ($g_q$) of next edge pixel $q$ then the distance between $p$ and $q$ is the stroke width else the ray tracing $p$ and $q$ is discarded. The pixels having similar stroke widths are grouped using connected component analysis. The letter candidates are chosen after some post processing based on the stroke width variance and aspect ratio. The letter candidates are grouped to give text regions. The idea is to segregate text from other high freuency content that might be present in the scene e.g. trees branches etc. We perform a bottom-up aggregation by merging pixels with similar stroke widths into connected components which allows in detecting characters across wide range of scales. It is able to identify near-horizontal text candidates. 

Stroke width of a region ($r$) is defined as \cite{li2014characterness},
\begin{equation}
SW(r)=\frac{SW_{var}(l)}{Mean(l)^2}
\end{equation}
where $l$ defines the shortest path between every pixel $p$ in the skeletal image of region ($r$) to the boundary of the region, $SW_{var}$ is the stroke width variance and $Mean$ gives the stroke width mean. 
We utilize the stroke width variance only which should be less for text candidates. We also store the values of stroke width as $\frac{SW_{mode}}{\sqrt{HXW}}$ and $\frac{SW_{max}-SW_{min}}{\sqrt{HXW}}$ (stroke width deviation)  where H and W denote height and width of the common region respectively. 
\item HOG and PHOG: PHOG consists of a histogram of orientation gradients over every sub-region in the image for every resolution level. The HOG vectors computed over each pyramid in the grid cells are concatenated. As compared to HOG, PHOG is more efficient. HOG is invariant to geometric and photometric transformations. In addition to this, PHOG helps in providing a spatial layout for the local shape of the image. Therefore, we utilize their combination as a characterness cue. 
\item Entropy: We calculate the entropy as the Shannon's entropy for the common regions ($b_{i} \cap R$) given as,
\begin{equation}
H=-\sum_{i=1}^{N-1}p_ilog(p_i)
\end{equation}
where $N$ denotes the number of gray levels and $p_i$ refers to the probability associated to the gray level $i$. In information
theory, entropy is the measure of average information
of a signal given its probability distribution. Higher entropy indicates higher disorder. In our scenario, text candidates shows lower variation in color values, thus typically there
is a dominating color in histogram having one sharp peak. However, for non-character candidates,
its color values span the histogram as result of
color variation. This corresponds to the entropy of the text candidates yielding smaller values than that of the non-text candidates and hence acts as an important cue in distinguishing among them and rejecting non-text candidates as described in the next section.
\end{enumerate}

\subsection{Bounding Box Refinement}
\begin{figure*}[ht]
\centering
\fbox{
\includegraphics[scale=0.5]{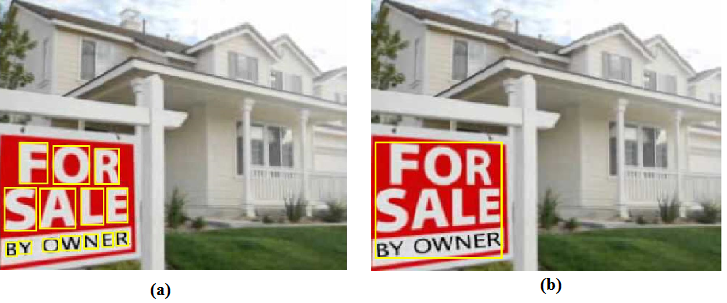}}
\caption{(a) Smaller regions in the blobs detected by eMSERs (b) Final result after postprocessing.}
\label{fig:postprocess}
\end{figure*}

The remaining set of regions are refined by calculating a set of parameters as stroke width distribution, pretrained characterness cues distribution and stroke width difference. We define a character cue distribution by computing the characterness cue values on ICDAR 2013 dataset. Additionally, we use this distribution to combine the neighboring candidate regions and aggregate them into one larger text region. We recompute the neighbors if they have similar distribution and reject otherwise. Finally, we combine all the neighboring regions into a single text candidate. Fig. \ref{fig:postprocess} shows the results of this post processing step.

\section{Experimental Results and Discussions}\label{sec:results}

\subsection{Experimental Setup and Datasets}
The experiments were performed on a $32$ GB RAM machine with Xeon $1650$ processor and $1$GB NVIDIA Graphics Card. Matlab $2015$b was used as the programming platform. The datasets used for evaluation of the proposed methodology are publicly available text datasets: MSRATD500 \cite{yao2012detecting} and KAIST \cite{lee2010scene}. MSRATD500 consists of $500$ images (indoor and outdoor scenes). The standard size of image varies between $1296$x$864$ to $1920$x$1280$. It consists of scenes capturing signboards with text in Chinese, English and mixed. The diversity and complex background in the images makes the dataset challenging. The KAIST scene text dataset consists of $3000$ images captured in different environmental settings (indoor and outdoor) with varying lighting conditions. The images are of size $640$x$480$. It consists of scenes with English, Korean and mixed texts. The majority of scenes are of shop and street numbers.
\subsection{Evaluation Methodology}

\subsubsection{Metrics.} The proposed technique is evaluated with precision, recall and F-measure metrics on the chosen datasets. The input for computing these metrics, is Intersection over Union (IoU) score, given as 
\begin{equation}
IoU=\frac{|S_1 \cap S_2|}{|S_1 \cup S_2|}
\end{equation}
where $S_1$ indicates the set of white pixels inside the blobs detected by our strategy before the elimination step (smaller individual blobs), $S_2$ indicates the set of white pixels inside the ground truth region and $\left |. \right |$ is the cardinality. The performance metrics in this paper are reported on blobs with majority of region being text i.e. having IoU $>0.5$.

\subsubsection{Training and Testing.} We perform training on ICDAR $2013$ \cite{karatzas2013icdar} dataset while the test set consists of MSRATD and KAIST datasets. This is unlike earlier methods where, in general, the training and testing samples are drawn from the same dataset. Moreover, such a setting makes the evaluation potentially challenging as well as allows us to evaluate the generalization ability of various techniques. The results on Characterness \cite{li2014characterness} and Blob Detection \cite{jahangiri2009attention} methods with training and testing sets as described earlier are reported using the publicly available source code. 

\subsection{Results}

\subsubsection{Qualitative Results.}
\begin{figure*}[thbtp]
\centering
\fbox{
\includegraphics[scale=0.35]{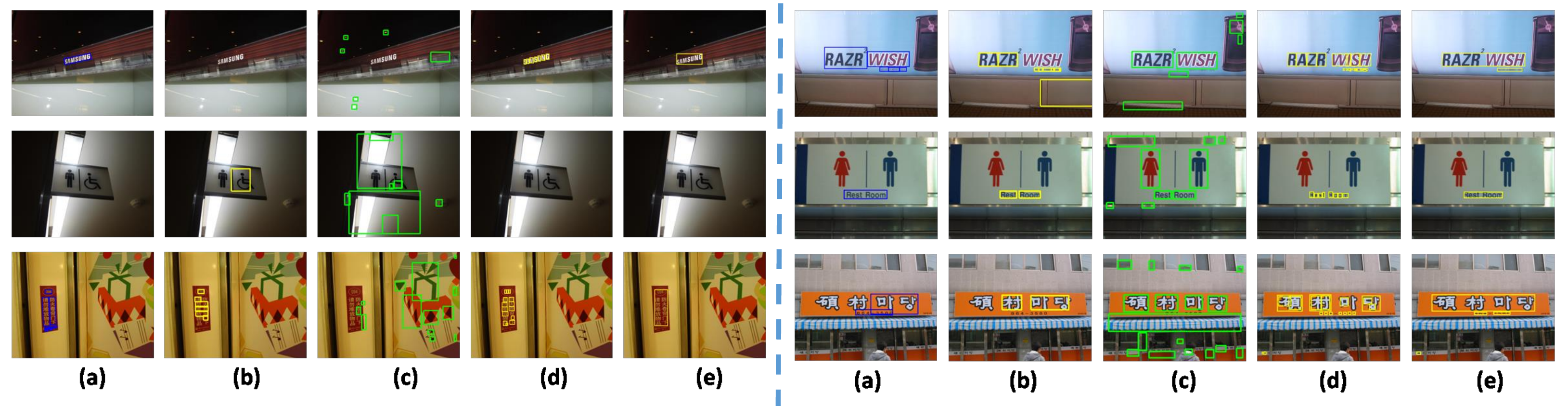}}
\caption{Results on i) MSRATD and ii) KAIST dataset images: (a) Ground Truth (b) Characterness \cite{li2014characterness} (c) Blob detection \cite{jahangiri2009attention} (d) Proposed Approach (before refinement step) (e) Proposed Approach }
\label{fig:msratd}
\end{figure*}

\begin{table*}[]
\footnotesize
\centering
\caption{Performance measures on MSRATD dataset}
\label{MSRA dataset}
\begin{tabular}{@{}lrrr@{}}
\toprule
\textbf{Method/Metric} & \multicolumn{1}{l}{\textbf{Precision}} & \multicolumn{1}{l}{\textbf{Recall}} & \multicolumn{1}{l}{\textbf{F-measure}} \\ \midrule
Proposed               & 0.85                                 & 0.33                              & 0.46                                 \\
Characterness \cite{li2014characterness}          & 0.53                                 & 0.25                              & 0.31                                 \\
Blob Detection \cite{jahangiri2009attention}         & 0.80                                 & 0.47                              & 0.55                                 \\
Epshtein et al. \cite{epshtein2010detecting}         & 0.25                                   & 0.25                                & 0.25                                   \\
Chen et al. \cite{chen2004detecting}         & 0.05                                   & 0.05                                & 0.05                            \\
TD-ICDAR \cite{yao2012detecting}              & 0.53                                   & 0.52                                & 0.50                                   \\
Gomez et. al. \cite{gomez2013multi}         & 0.58                                   & 0.54                                & 0.56                                   \\ \bottomrule
\end{tabular}
\end{table*}
Figure \ref{fig:msratd} shows qualitative results on a few example images from MSRATD and KAIST datasets. It can be observed that the images obtained after region refinement demonstrate better localization of textual regions while those on MSRATD dataset (Fig. \ref{fig:msratd} (i)) show tighter localization as compared to other techniques. One of the aims of the proposed technique is to reduce false positives, which can be observed from the second row of Fig. \ref{fig:msratd} (i) where the proposed method provides a tight bounding box on text regions while there are false positives with other techniques except Characterness. The signboard in the image does not consist of any text data still the contemporary methods detect it as a text candidate. This could be due the fact that the signboard consists of a rounded sketch which may correspond to alphabets such as 'O', 'Q' etc. Since the proposed technique strictly encodes the stroke width variance along with other characterness cues, we are able to avoid detection of such false candidates. Similar findings are observed for the KAIST dataset as well.

\subsubsection{Quantitative Results.}

\begin{table*}[]
\footnotesize
\centering
\caption{Performance measures on KAIST dataset}
\label{KAIST dataset}
\begin{tabular}{llll}
\hline
\multicolumn{4}{c}{\textbf{KAIST-English}}                                                         \\ \hline
\textbf{Method/Metric}    & \textbf{Precision} & \textbf{Recall} & \textbf{F-Measure} \\ \hline
Proposed & 0.8485                 & 0.3299                  & 0.4562                     \\ 
Characterness    & 0.5299                 & 0.2476                  & 0.3136                     \\ 
Blob Detection & 0.8047                 & 0.4716                  & 0.5547                     \\ \hline
\multicolumn{4}{c}{\textbf{KAIST-Korean}}                                                          \\ \hline
\textbf{Method/Metric}    & \textbf{Precision} & \textbf{Recall} & \textbf{F-measure} \\ \hline
Proposed & 0.9545                 & 0.3556                  & 0.4994                     \\ 
Characterness    & 0.7263                 & 0.3209                  & 0.4083                     \\ 
Blob Detection \cite{jahangiri2009attention} & 0.9091                 & 0.5141                  & 0.6269                     \\ \hline
\multicolumn{4}{c}{\textbf{KAIST-Mixed}}                                                           \\ \hline
\textbf{Method/Metric}    & \textbf{Precision} & \textbf{Recall} & \textbf{F-measure} \\ \hline
Proposed & 0.9702                 & 0.3362                  & 0.4838                     \\
Characterness    & 0.8345                 & 0.3043                  & 0.4053                     \\ 
Blob Detection & 0.9218                 & 0.4826                  & 0.5985                     \\ \hline
\multicolumn{4}{c}{\textbf{KAIST-All}}                                                           \\ \hline
\textbf{Method/Metric}    & \textbf{Precision} & \textbf{Recall} & \textbf{F-measure} \\ \hline
Proposed & 0.9244                 & 0.3407                  & 0.4798                     \\
Characterness \cite{li2014characterness}   & 0.6969                 & 0.2910                  & 0.3757                     \\ 
Blob Detection \cite{jahangiri2009attention} & 0.8785                 & 0.4894                  & 0.5933                     \\ 
Gomaz et al. \cite{gomez2013multi}  & 0.66 & 0.78 & 0.71 \\
Lee et al. \cite{lee2010scene}  & 0.69 & 0.60 & 0.64 \\
\hline
\end{tabular}
\end{table*}

Table \ref{MSRA dataset} and \ref{KAIST dataset} show empirical results on MSRA and KAIST scene datasets respectively. From the empirical results, it can be seen that on MSRATD dataset, the proposed method achieves significantly higher precision and F-measure as compared to Characterness while having a $28$\% (precision) and $64$\% (F-measure) gain and and a slightly lower ($\sim6$\%) recall rate with Blob Detection. The proposed technique outperforms the compared methods on precision while performs close in terms of F-measure and recall. It is important to note here that the proposed technique does not involve any explicit training allowing the technique to be directly extensible to domains such as symbol identification, road sign identification etc. On KAIST dataset, the proposed method consistently outperforms Characterness on all benchmarks with average improvement of $36$\%, $17$\% and $29$\% in precision, recall and F-measure respectively. The proposed technique also achieves better precision as compared to Blob Detection. The results show that the proposed method is able to generalize better on a test set while being trained on an entirely distinctive character set. For completeness in comparison, we also provide performance of other techniques on KAIST dataset. However, it should be noted that the objective of these techniques is generally to maximize text detection specifically for a script or to attain script independence with curated training examples with the mixture of scripts to be detected. This possibly makes the comparison with proposed technique tougher as the objective is to obtain better generalization ability. 


\section{Conclusion}\label{sec:conclusion}
This paper proposed an effective text detection scheme by utilizing stronger characterness measure. A post processing step is used to reject the non-textual blobs and combine smaller blobs obtained by eMSERs into one larger region. The effectiveness of  the proposed scheme has been analyzed with precision, recall and F-measure evaluation measures showing that the proposed scheme performs better than the traditional text detection schemes.

%
%
\bibliographystyle{splncs03} 
\bibliography{ref}

\end{document}